\documentclass[letterpaper]{article} 
\usepackage{aaai24}  
\usepackage{times}  
\usepackage{helvet}  
\usepackage{courier}  
\usepackage[hyphens]{url}  
\usepackage{graphicx} 
\usepackage{natbib}  
\usepackage{caption} 
\usepackage{algorithm}
\usepackage{algorithmic}
\usepackage{amsfonts}
\usepackage{amsmath}
\usepackage{subfig}
\usepackage{multirow}
\usepackage{colortbl}
\usepackage{xcolor}
\newcommand{\mc}{\mathcal}
\newcommand{\bb}{\mathbb}

\graphicspath{ {./images/}}
\usepackage{newfloat}
\usepackage{listings}
\DeclareCaptionStyle{ruled}{labelfont=normalfont,labelsep=colon,strut=off} 
\floatstyle{ruled}
\newfloat{listing}{tb}{lst}{}
\floatname{listing}{Listing}
\title{RLPeri: Accelerating  Visual Perimetry Test with Reinforcement Learning and Convolutional Feature Extraction}
\author{
    Tanvi Verma\textsuperscript{\rm 1},
    Linh Le Dinh\textsuperscript{\rm 1},
    Nicholas Tan\textsuperscript{\rm 2},
    Xinxing Xu\textsuperscript{\rm 1},
    Chingyu Cheng\textsuperscript{\rm 2},
    Yong Liu\textsuperscript{\rm 1}
}
\affiliations{
    \textsuperscript{\rm 1}Institute of High Performance Computing (IHPC), Agency for Science, Technology and Research (A*STAR), Singapore\\
    \textsuperscript{\rm 2}Singapore Eye Research Institute, Singapore National Eye Centre, Singapore\\
    \{vermat,linhld,xuxinx,liuyong\}@ihpc.a-star.edu.sg, yqntan@gmail.com, chingyu.cheng@duke-nus.edu.sg

}

\usepackage{bibentry}

\begin{document}

\maketitle

\begin{abstract}
Visual perimetry is an important eye examination that helps detect vision problems caused by ocular or neurological conditions. During the test, a patient's gaze is fixed at a specific location while light stimuli of varying intensities are presented in central and peripheral vision. Based on the patient's responses to the stimuli, the visual field mapping and sensitivity are determined. However, maintaining high levels of concentration throughout the test can be challenging for patients, leading to increased examination times and decreased accuracy.

In this work, we present RLPeri, a reinforcement learning-based approach to optimize visual perimetry testing. By determining the optimal sequence of locations and initial stimulus values, we aim to reduce the examination time without compromising accuracy. Additionally, we incorporate reward shaping techniques to further improve the testing performance. To monitor the patient's responses over time during testing, we represent the test's state as a pair of 3D matrices. We apply two different convolutional kernels to extract spatial features across locations as well as features across different stimulus values for each location. Through experiments, we demonstrate that our approach results in a 10-20\% reduction in examination time while maintaining the accuracy as compared to state-of-the-art methods. With the presented approach, we aim to make visual perimetry testing more efficient and patient-friendly, while still providing accurate results.
\end{abstract}

\section{Introduction}\label{sec:intro}
Perimetry, also known as the visual field test, is a crucial diagnostic tool used to assess a patient's visual abilities. The test measures the extent of the visual field (VF), and the sensitivity of vision in different areas without the need for eye movements. It is an essential examination for identifying and monitoring a range of ocular and neurological conditions, such as glaucoma, which initially affects peripheral vision and can progress to total blindness if left untreated. With an estimated 80 million people suffering from glaucoma worldwide in 2020, and a projected increase to 111.8 million by 2040 \citep{tham2014global}, the importance of improving the perimetry test procedure cannot be overstated. The development of innovative methods that make perimetry testing more efficient and effective will benefit both patients and society as a whole.

Perimetry tests require patients to maintain focus on a fixed reference point while pressing a button in response to presented light stimuli of varying intensities. The test results are used to determine the sensitivity threshold at each location, and the examination time can range from 3 minutes to 15 minutes per eye, depending on the type of test and the condition being diagnosed \citep{bengtsson1998evaluation,chauhan1999test,phu2021viability}. For older patients, who are at a higher risk of developing glaucoma, the prolonged examination time can be taxing and potentially impact the accuracy of the test results. The perimetry test heavily relies on patient attention, and prolonged testing times can decrease the reliability and accuracy of the results. Hence, there is a need to optimize the perimetry procedure to reduce examination times while maintaining test accuracy. By improving the efficiency of the perimetry test, patients will benefit from a more streamlined and less overwhelming diagnostic experience.

\begin{figure}
  \centering
  {\includegraphics[scale=0.25]{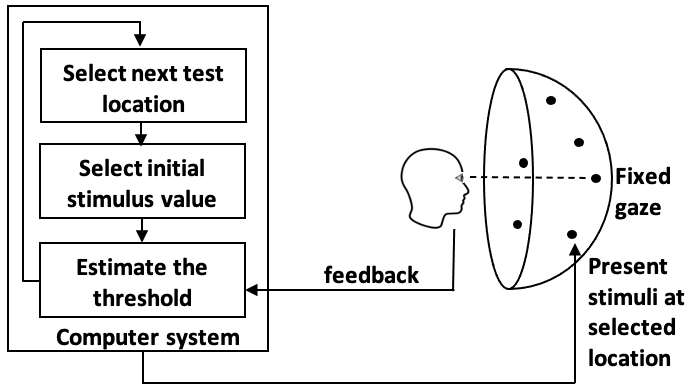}}
  \caption{Three decision-making steps of visual field testing. }
  \label{fig:vf-framework}
\end{figure}
As shown in Figure \ref{fig:vf-framework}, the perimetry test can be divided into three critical decision-making steps: (1) selection of the next testing location, (2) determination of the initial stimulus intensity to be presented at the next location, and (3) estimation of the sensitivity threshold at that location. This sequential decision-making problem is characterized by uncertainty, as the patient's response to the presented stimuli is probabilistic. Reinforcement learning (RL) presents a suitable approach for optimizing the perimetry test, as it can determine the optimal sequence of locations and initial stimulus values to be tested to maximize speed and accuracy. In this paper, we propose RLPeri, an RL-based perimetry strategy, which focuses on optimizing the entire visual field testing procedure. For the third step, we use Zippy Estimation by Sequential Testing (ZEST) \citep{king1994efficient} method to estimate the sensitivity threshold at each location. Our approach differs from recent works, such as Patient-Adaptive Sampling Strategy (PASS) \citep{kucur2019patient}, which only selects a predefined number of locations to be tested. Our method optimizes the entire visual field testing sequence and shows improved test accuracy compared to PASS. Our proposed method is designed to address the requirement for higher accuracy in clinics. In this particular case, selecting fewer locations like PASS is insufficient to meet this requirement.

Since the 1970s, perimetry tests has been automated by computer algorithms \cite{johnson2011history}. These algorithms have a trade-off between accuracy versus speed of assessment. A detailed assessment aiming for high precision and accuracy comes at the price of long testing times, patient fatigue, and consequently (and paradoxically) less reliable patient results. On the other hand, a cursory/fast assessment of the visual field may produce too imprecise/inaccurate results.

To address this trade-off between speed and accuracy, previous research has either proposed reducing the testing duration by testing only a selected number of locations \citep{kucur2017sequentially,kucur2019patient} or predicting the sensitivity threshold values \citep{shon2022can,park2019visual}. In our work, we adopt a unique approach by using potential-based reward shaping to strike a balance between speed and accuracy. The potential, or reconstruction error, is calculated as the difference between the suggested initial value and the actual ground truth sensitivity value and used to shape the reward. Furthermore, in order to capture the connection between the presented stimulus at a specific location and the corresponding feedback from the patient, along with sensitivity values from neighboring locations, we represent the test state using a pair of 3D matrices. Subsequently, we employ group-wise \citep{krizhevsky2012imagenet} and point-wise \cite{howard2017mobilenets} convolution kernels to extract the relevant features from the matrices. Group-wise kernels have been demonstrated to efficiently extract features in a variety of scenarios \cite{douillard2020podnet,verma2021efficient}. In our work, this approach enables us to examine the relationship between stimuli, patient feedback, and sensitivity values across neighboring locations effectively. To summarise, following is our main contribution
\begin{itemize}
    \item We formulate perimetry testing as a Markov Decision Process (MDP) and propose RLPeri method to determine the optimal sequence of locations and initial testing intensities for faster and more accurate results.
    \item We address the trade-off between speed and accuracy by utilizing potential based reward shaping. This technique uses reconstruction error as the potential of the current state, helping to achieve a balanced outcome.
    \item We represent test's state as a pair of 3D matrices, and to extract features related to presented stimulus values and threshold values of neighboring locations, we utilize two distinct convolution kernels. 
    \item Through rigorous experimentation, we demonstrate that our RLPeri approach outperforms existing state-of-the-art methods in terms of speed while maintaining the accuracy.
\end{itemize}
\begin{figure*}
  \centering
  \subfloat[\label{fig:sensitivity}]{\includegraphics[scale=0.38]{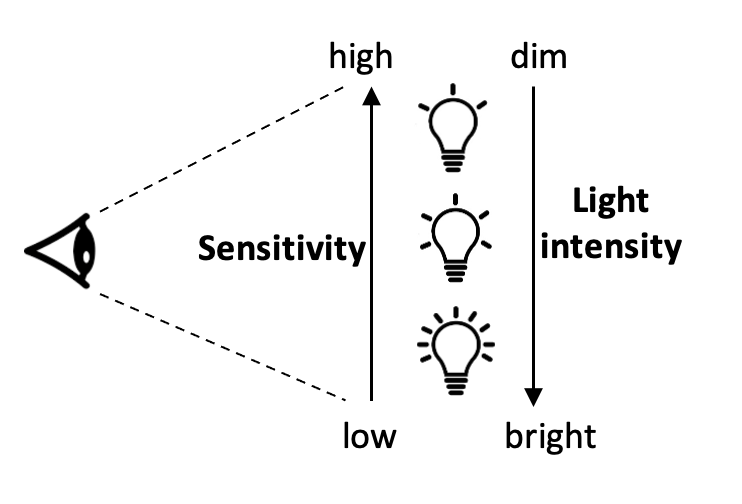}} 
  \subfloat[\label{fig:fos}]{\includegraphics[scale=0.35]{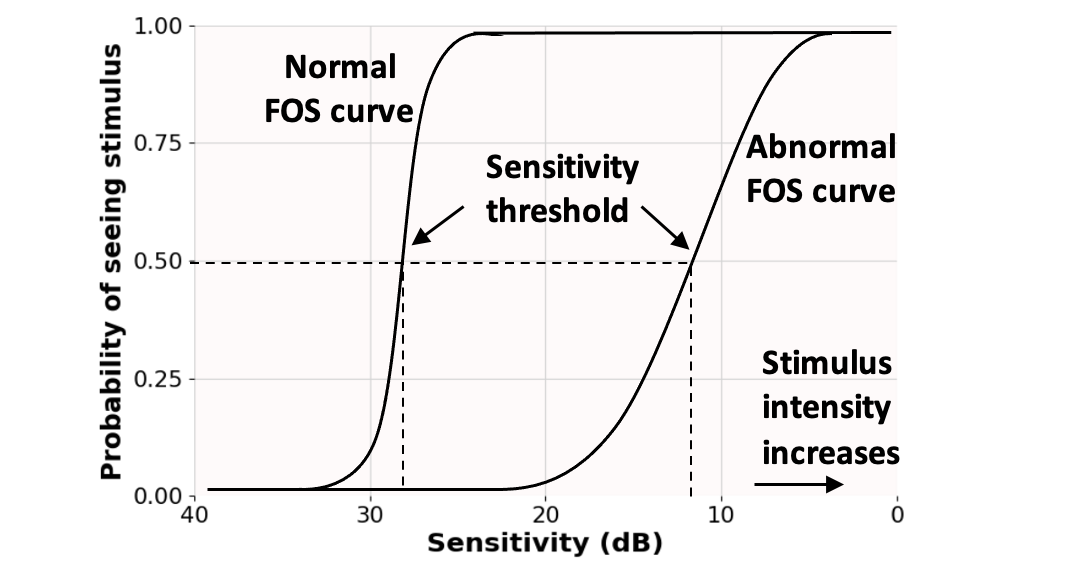}}
  \subfloat[\label{fig:vf}]{\includegraphics[scale=0.35]{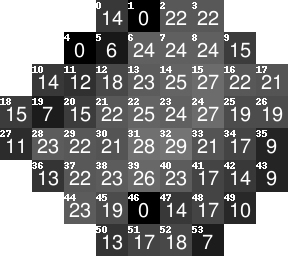}}
  \caption{\protect\subref{fig:sensitivity} The inverse relationship between light intensity and sensitivity to light. Individuals with a high sensitivity to light are able to detect even faint lights, while those with low sensitivity to light can only see very bright lights. \protect\subref{fig:fos} Frequency-of-seeing (FOS) curve. Sensitivity threshold is defined as the intensity at which probability of seeing is 50\%. Note the inverse relationship between sensitivity and stimulus intensity. \protect\subref{fig:vf} 2D mapping of sensitivity threshold values for 54 locations of 24-2 test pattern. Darker color indicates low sensitivity.}
\end{figure*}
\section{Background} \label{sec:bg}
In this section we briefly discuss about perimetry, ZEST procedure and reinforcement learning.
\subsection{Perimetry}
Perimetry testing is a crucial procedure in ophthalmology to assess a patient's visual field and sensitivity thresholds. The test is done by presenting light stimuli of different intensities (in decibels, dB) at different locations of visual field. The relationship between stimulus intensity and a patient's sensitivity to light is inverse. As shown in Figure \ref{fig:sensitivity}, a patient with a high sensitivity to light can see with a low-intensity stimulus, whereas a patient with low sensitivity to light requires a high-intensity stimulus. The sensitivity to light is measured in decibels (dB) as follows
\footnotesize
\begin{align}
  dB = 10 * log(L_{max}/L)  \label{eq:db}
\end{align}
\normalsize
Where $L_{max}$ is the maximum luminance (intensity of light reflected on the perimetric surface, measured in apostilb, asb) the perimeter can display and $L$ is luminance of the stimulus at the threshold \cite{racette2018visual}. 0 dB represents the highest level of intensity that the perimetry equipment can produce. 

Following stimulus presentation, the patient responds \textit{seen}/\textit{not seen} depending on their ability to perceive the stimulus. Based on their response, the intensity of the stimuli is adjusted until the sensitivity threshold is estimated for that specific location. The procedure is then repeated again for the next location till the sensitivity threshold is estimated for all the visual field locations. However, it is a noisy process and same patient always show slightly varying responses if test is repeated. In short, the likelihood of \textit{seeing} or \textit{not seeing} is probabilistic, especially for stimuli values near threshold value. The sensitivity threshold is defined as the intensity of light that can be perceived by the patient 50\% of the time. Figure \ref{fig:fos} depicts frequency-of-seeing (FOS) curve, which describes the probability that a patient will see a target as a function of the intensity of the stimulus. The FOS curve on the left with a threshold value of 32 dB represents a normal eye, while the FOS curve on the right with a lower threshold value indicates some visual defect. A sensitivity threshold of 0 dB means that the patient is not able to see the most intense stimulus that the perimetry equipment can produce. 

The number of locations in the visual field varies depending on the test pattern, for example a 24-2 test strategy \citep{humphry1993vf} test 54 locations. Figure \ref{fig:vf} illustrates a 2D representation of visual fields specific to the 24-2 test strategy. The visualization includes sensitivity threshold values (measured in dB) at the center and location identifiers placed in the upper left corner. The threshold values generally varies from 0-40 dB.

\subsection{ZEST Procedure} 
The ZEST \citep{king1994efficient} procedure involves associating each visual field location with a probability density function (pdf), which characterizes the probability that a specific stimulus value is the threshold for that location. Initial pdf, a likelihood function and a stopping criteria are key components of ZEST. The test at a given location begins with an initial pdf. Subsequently, the procedure iteratively refines the pdf by incorporating the patient's responses and the likelihood function. The test at that location concludes when the stipulated stopping criterion is met. One possible way to stop the test is to terminate it when the standard deviation of the pdf is lower than a predetermined value \cite{turpin2002development}, which we used in our experiments.

\subsection{Reinforcement Learning}
Reinforcement Learning \citep{sutton1998reinforcement} is a popular method for solving Markov Decision Process (MDP) when the model of MDP is not known. An MDP is formally defined as the tuple $\big \langle {\mc{S,A,T,R }}\big \rangle$, where $\mc{S}$ is the set of states, $ \mc{A}$ is the set of actions, $\mc{T}(s,a,s')$ represents the probability of transitioning from state $s$ to state $s'$ on taking action $a$ and $\mc{R}(s,a)$ represents the reward obtained on taking action $a$ in state $s$. RL agents learn a policy that maximizes their expected future reward while interacting with the environment. Q-learning \citep{watkins1992q} is one of the most popular RL approach, where the action value function $Q(s,a)$ are updated based on experiences given by $(s,a,s',r)$:
\footnotesize
\begin{align}
{Q}(s,a) \leftarrow  (1-\alpha){Q}(s,a) + \alpha [r + \gamma \max_{a'} Q (s',a')] \label{eq:qval}
\end{align} 
\normalsize
Where $\alpha$ is the learning rate and $\gamma$ is the discount factor. 
Advantage function $A(s,a)$ is a measure of how much action $a$ is good or bad decision in state $s$, in short it measures the advantage of selecting action $a$
\footnotesize
\begin{align}
    A(s,a) = Q(s,a) - V(s)
\end{align}
\normalsize
where $V(s)$ is the state value function.
DQN \citep{mnih2015human} approximates the Q-values with a deep neural network. This deep network for Q-values is parameterized by a set of parameters, $\theta$ and the parameters are learned using an iterative approach that employs gradient descent on the loss function. Specifically, the loss function at each iteration is defined as follows:
\footnotesize
\begin{align}
\mathcal{L}_{\theta} = \mathbb{E} [(y^{DQN} - Q(s,a;\theta))^2] \label{q-loss}
\end{align}
\normalsize
where $y^{DQN} = r + \gamma \text{max}_{a'} Q(s', a'; \theta^-)$ is the target value computed by a target network parameterized by previous set of parameters, $\theta^-$. Parameters $\theta^-$ are frozen for some time while updating the current parameters $\theta$. To ensure independence across experiences this approach maintains a replay memory ${\mathcal J}$ and then samples experiences from it. 
\subsection{Knowledge Based Learning}
Knowledge based reinforcement learning deals with incorporating domain knowledge into the learning to guide exploration. Reward shaping is one of such approach which provides additional reward signal representative of prior knowledge in additional to the reward received from the environment. Reward $r$ is augmented with additional reward signal $F(s,s')$, which is a general form of any state based reward shaping function. Reward shaping has been shown to improve learning \citep{ng1999theory}. However it can result into agent learning unexpected/undesired behavior if used improperly. For example, \citet{randlov1998learning} demonstrated that by providing additional reward to stay balanced while learning to ride a bicycle resulted into learning a policy which preferred staying balanced and cycling in circles than reaching the destination.

To avoid such behavior, \citet{ng1999policy} proposed potential based reward shaping, where additional reward is computed as the difference in some potential function $\varphi$ defined over states. The potential based reward is computed as 
\footnotesize
\begin{align}
F(s, s') = \gamma \varphi(s') - \varphi(s) \label{eq:pot}
\end{align}
\normalsize
Where $\gamma$ is the same discount factor used in update rule provided in Equation \ref{eq:qval}. The potential of a state represents the goodness of the state, for example, potentials of states close to goal state are generally set to be more than the states that are far from the goal state. Potential based reward shaping algorithms do not alter the optimal policy. 

\section{RLPeri: Reinforcement Learning Based Perimetry} \label{sec:rlp}
\begin{figure}
  \centering
  {\includegraphics[scale=0.45]{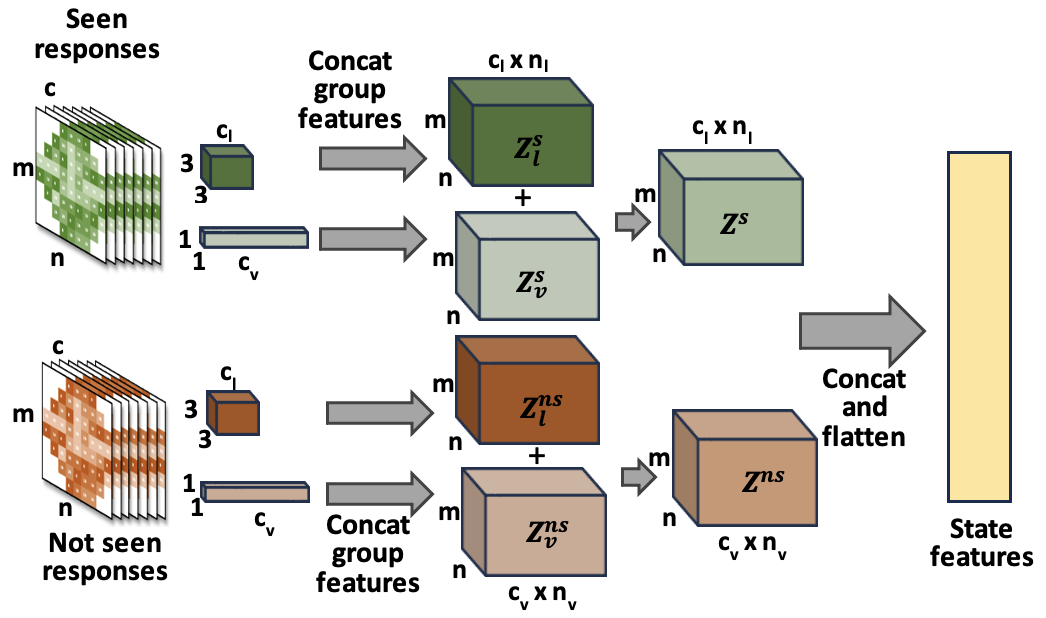}} 
  \caption{The test's status is depicted using a pair of 3D matrices, encompassing the patient's responses. Each matrix is designated for \textit{seen} responses and \textit{not seen} responses, respectively. Group-wise and point-wise convolutional kernels are employed to capture features related to the threshold values of the surrounding locations as well as the patient's responses to different stimulus values presented at a given location.}
  \label{fig:feat}
\end{figure}
As discussed in previous section, MDP is a mathematical framework that models decision making problems. In the context of visual field testing, the MDP represents the process of performing a visual field test on a patient.

\subsection{State Space and State Features} 
The state space of the MDP, $\mathcal{S}$, consists of all the possible visual fields that are represented as a pair of 3D matrix of size $c \times m \times n$. Here $m$ and $n$ are dependent upon the specific test pattern. For instance, in the case of the 24-2 test pattern, a total of 54 locations can be represented using an $8 \times 9$ 2D matrix, with corner locations masked as demonstrated in Figure \ref{fig:vf}. The variable $c$ denotes the number of possible stimulus values.
Consequently, the state can be conceptualized as a collection of $c$ 2D matrices, each dedicated to a distinct stimulus value. First matrix compiles all the \textit{seen} responses, while the second matrix accounts for all the \textit{not seen} responses of the patient during testing. The entry at position $[i, j, k]$ ($i \in [1, c]$, $j \in [1, m]$, $k \in [1, n]$) of the \textit{seen} matrix represents the number of times the patient was able to perceive a stimulus with an intensity of $i$ dB at location $[j, k]$. Similarly, the entry at position $[i, j, k]$ in the \textit{not seen} matrix represents the number of times the patient fails to perceive the stimulus with an intensity of $i$ dB at location $[j, k]$. As discussed earlier, the patient's ability to perceive a stimulus is probabilistic. Consequently, the algorithm may repeatedly present the same stimulus value at a chosen location. This is the reason we maintain a count of both \textit{seen} and \textit{not seen} responses for each stimulus value. The purpose of employing this state representation is to comprehensively capture the entirety of the patient's responses, enabling subsequent extraction of pertinent features for the purpose of learning the state values.

Figure \ref{fig:feat} depicts the process of extracting the state feature. We use two types of group-wise convolutional kernels, $k^l \in \bb{R}^{3 \times 3 \times c_l}$ and $K^v \in \bb{R}^{1 \times 1 \times c_v}$. $c_l$ and $c_v$ ($c_l, c_v < c $) are number of channels in each group and $n_l$ and $n_v$ are number of groups in $K^l$ and $K^v$ respectively. We ensure\footnote{As $c=41$ if we consider a range of 0-40 dB, we used a convolution layer to increase the number of channels of the original input to 64.} that $c_l \times n_l = c_v \times n_v = c$. $K^l$ is utilized to capture spatial characteristics based on threshold values of neighboring locations, as these sensitivity thresholds often exhibit correlations. Similarly, to capture features across various stimulus values presented at a location, we employ $K^v$, which is a pointwise kernel. Since the patient's ability to perceive a presented stimulus is probabilistic, if the patient is unable to perceive almost all of the consecutive stimuli in a selected stimulus intensity range, it is likely that the threshold value at that location is lower than the selected range. Conversely, if the patient successfully perceives the majority of stimuli within the selected range, it suggests that the actual threshold is likely situated above that range. This is why we use a pointwise kernel to help the model decide on the starting stimulus value. After extracting group-wise features, these features are combined to yield $Z^s_l$ and $Z^s_v$ features from the \textit{seen} matrix, as well as $Z^{ns}_l$ and $Z^{ns}_v$ features from the \textit{not seen} matrix. These features are then added to obtain the respective $Z^s$ and $Z^{ns}$ features, which are subsequently concatenated and flattened to produce the final state features.

\subsection{Action Space} 
The set of possible visual field locations is represented by $L$, and the set of possible stimulus values is represented by $V, |V|=c$. The action taken in the MDP is represented as a tuple $(a_l, a_v)$, where $a_l \in L$ represents the location in the visual field where the stimulus is to be presented and $a_v \in V$ represents the stimulus value in dB. The state transition probability depends on the FOS curve for the selected location, which models the patient's probabilistic response to the stimulus.

\subsection{Reward Function} 
The reward function $\mathcal{R}(s, a_l, a_v)$ represents the negative of the number of stimuli presented to complete the test at location $a_l$ in the current state $s$, given that the initial stimulus value presented was $a_v$. To reconcile the conflicting objectives of speed (the number of stimuli presented) and accuracy (the mean squared error between the ground truth values and the constructed visual field), we use the potential based reward shaping. The potential $\varphi(s)$ of the state is represented by the negative mean squared error between the already discovered locations in the current state and their corresponding ground truth values. Consequently, during the target computation in Equation \ref{target}, $r$ is replaced with $r + F(s,s')$, where $F(s, s')$ is computed as outlined in Equation \ref{eq:pot}. 
 
\begin{figure}
  \centering
  {\includegraphics[scale=0.36]{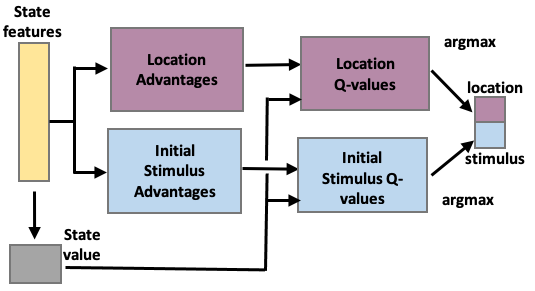}} 
  \caption{Framework for RLPeri. Q-values are learned independently for each action dimension.}
  \label{fig:arch}
\end{figure}

The action space of the MDP is huge due to its two dimensions. Hence, to tackle large action space, we use Branching Dueling Q-Network (BDQ) \citep{tavakoli2018action} architecture to optimize learning. Figure \ref{fig:arch} shows the framework of RLPeri where $Q$-values of each action dimension are learned separately and independently, allowing for more effective and efficient learning. 
The shared state representation allows for estimation of both the state values and advantages of each action dimension, resulting in improved overall performance. We define $Q_l(s,a;\theta)$ and $Q_v(s,a;\theta)$ as the parameterised action-value functions for location dimension and stimuli value dimension actions. Here $\theta$ is the network parameter. These values are estimated using the state value function $V(s, \theta)$ and the advantage functions $A(s, a_l, \theta)$ and $A(s, a_v, \theta)$ as follows
\footnotesize
\begin{align*}
    Q_l(s,a;\theta) = V(s,\theta) + \Big( A(s,a_l;\theta) - \dfrac{1}{|L|} \sum_{a_l'} A(s,a_l';\theta) \Big) \\
    Q_v(s,a;\theta) = V(s,\theta) + \Big( A(s,a_v;\theta) - \dfrac{1}{|V|} \sum_{a_v'} A(s,a_v';\theta) \Big)
\end{align*} 
\normalsize

After estimating Q-values for each action dimension, the target values are computed by averaging the next state's action-values for each dimension. For experience $<s, a_l, a_v, r, s'>$, the target value is computed as follows

\footnotesize
\begin{align}
y = r + & \gamma \dfrac{1}{2} \Big[Q_l\Big(s',\mathrm{argmax}_{a'_l} Q_l(s', a'_l; \theta); \theta^-\Big) \nonumber \\ &+ Q_v\Big(s', \mathrm{argmax}_{a'_v} Q_v(s', a'_v; \theta); \theta^-\Big) \Big] \label{target}
\end{align}
\normalsize

Finally, the the network is trained on batch of experiences replayed from the replay memory ($\mc{J}$) by computing loss values for each action dimension. The loss function is defined as 
\footnotesize
\begin{align}
    \mc{L} = \bb{E}_{(s,a_l, a_v,r,s') \sim \mc{J}} \Big[\dfrac{1}{2}\big[ \big(y - Q_l(s,a_l; \theta)\big) \nonumber \\ + \big(y - Q_v(s,a_v; \theta)\big) \big] \Big]^2 \label{loss}
\end{align}
\normalsize

The training is episodic where one complete visual field test (i.e. all $L$ locations have been tested) is considered as one episode. RLPeri uses $\epsilon-$greedy exploration \citep{sutton1998reinforcement} to generate learning episodes. Experiences $<s, a_l, a_v, r, s'>$ are used to update the learning. After selecting next location $a_l$ and corresponding starting stimulus value $a_v$, ZEST is used to estimate the threshold value at the location. The detailed steps of RLPeri and ZEST are provided in the Appendix.

\section{Experiments}
In this section we discuss our experimental results in detail. We first describe our dataset and simulator. 
\subsection{Dataset}
We evaluated our approach on Rotterdam visual field data set \citep{erler2014optimizing}. It includes 5108 visual field samples from 22 healthy and 139 glaucomatous patients. Visual Fields were acquired using 24-2 pattern (54 locations). We divided the data set into training data (60\%), test data (20\%) and validation data (20\%).

\subsection{Simulator}
To train and evaluate RLPeri, we created a realistic simulator using the dataset. As mentioned, the patient responses are probabilistic, with the likelihood of a \textit{seen} or \textit{not seen} response being dependent on the location's FOS curve. The ground truth data only provides visual field maps with corresponding sensitivity threshold values, but does not include the actual FOS curve. To address this, we estimated the FOS curve for each location by fitting a normal cumulative distribution function to the ground truth threshold value, with a fixed standard deviation of 1. 

In RLPeri, each complete VF test is treated as a single learning episode. At the start of each episode, the initial state is represented with two 3D matrices of size $41 \times 8 \times 9$, one each for \textit{seen} and \textit{not seen} matrix. Non-VF locations (corner locations if we represent 2D map of Figure \ref{fig:vf} using $8 \times 9$ matrix) in the each 2D matrix are initialized to -2 while valid VF locations are initialized with a count of 0. During training and testing, the simulator presents a stimulus value to a location and simulates a \textit{seen} or \textit{not seen} response based on the estimated FOS curve and its corresponding probability distribution. In order to simulate the responses of \textit{seen} and \textit{not seen} at a particular location for a given stimulus, we generate a random number between 0 and 1. Let p represent the probability of seeing at that location for the presented stimulus, as determined by the FOS curve. If the randomly generated number is less than or equal to p, we assume that the stimulus was \textit{seen}. Otherwise, if the generated number is greater than p, we consider the stimulus to be \textit{not seen}. The sate moves to the next state based on these responses.

We employed a stimulus value range of 0 to 40 dB. After determining next location to be tested and initial stimulus value, we use ZEST to estimate the final sensitivity threshold value at the location to reconstruct the VF. As described in the background section, initial probability density function (pdf) for each location is needed. The training dataset was utilized to calculate the frequency distribution of sensitivity values at each location, which was subsequently employed to formulate the initial pdf for each location. Similarly, we assumed liklihood function to be a gaussian function with presented stimuli as mean and a fixed standard deviation of 0.5 dB. We used standard deviation (represented as $\sigma$ in experimental results) of the estimated pdf as the stopping criteria. When $\sigma$ of estimated pdf falls below a predefined value (we experimented with $\sigma=1,2,3$), the testing at the location is considered to be complete.

\subsection{Evaluation Metric}
We used two different metrics to compare the performances. First one is total number of stimuli presented to complete the visual field test. As time taken to complete the test is linearly correlated with total number of stimuli presented, we use total number of stimuli presented as the proxy of speed of the visual field test. Mean Squared Error (MSE) between the ground truth visual field and the reconstructed visual field is used as the second evaluation metric. As illustrated in Equation \ref{eq:db}, the correlation between sensitivity and light intensity is non-linear. Similarly, the relationship between visual function and light intensity exhibits non-linearity. For example, a 90 asb increase is noticeable when intensity goes from 10 to 100 asb, but the same increase is hardly noticeable when it goes from 1,000 to 1,090 asb \cite{racette2018visual}. This method of measuring sensitivity from light intensity effectively imparts linearity to the relationship between dB and intensity. Therefore, the visual effect difference between higher dB values (such as 24 dB and 30 dB) is similar to the visual effect difference between lower dB values (such as 2 dB and 8 dB). This characteristic makes MSE an appropriate metric for assessing accuracy.

The experiments were carried out across five separate seeds, and the results encompass both the reported mean values and their respective standard deviations. In the Appendix, we additionally conduct a qualitative analysis of the generated sequence of locations and initial stimulus values across multiple visual fields.

\subsection{Hyperparameter and Computing Infrastructure }
We used Adam optimizer \citep{kingma2014adam} to perform mini-batch stochastic gradient descent optimization. We also use dropout layer and layer normalization to prevent the network from overfitting. We set number of groups in $K^l$ to 8 and number of groups in the pointwise convolutional kernel $K^v$ to 4. We used 1e-4 as learning rate after tuning it in the range of 1e-2-1e-4. The batch size was set to 2048. We performed $\epsilon$-greedy exploration and it was decayed exponentially. We stopped decaying $\epsilon$ once its value reached to 0.01. We also use validation error while training to select the best model. Our experiments were conducted on a 64-bit Ubuntu machine equipped with 500GB of memory and a 40GB GPU. For deep learning tasks, we employed the PyTorch framework.

\subsection{Results}

\begin{table}
\centering
\begin{tabular}{| c | c | c | c |  }
\hline 
\multirow{2}{*}{$\sigma$} & \multirow{2}{*}{Method} & Number of & \multirow{2}{*}{MSE} \\ 
& & stimuli presented &\\ \hline
\multirow{3}{*}{1} & ZEST & 348.15 (0.66) & 0.949 (0.005)  \\ \arrayrulecolor{lightgray}\cline{2-4} \arrayrulecolor{black}
& SORS  & 364.62 (0.66) & 0.943 (0.007) \\ \arrayrulecolor{lightgray} \cline{2-4} \arrayrulecolor{black}
& RLPeri & \textbf{310.56} (3.2) & \textbf{0.938} (0.013)  \\  \arrayrulecolor{black}\hline
\multirow{3}{*}{2} & ZEST  & 278.68 (0.54) & 1.475 (0.01) \\ \arrayrulecolor{lightgray} \cline{2-4} \arrayrulecolor{black}
& SORS  & 295.93 (0.36) & 1.454 (0.009) \\ \arrayrulecolor{lightgray} \cline{2-4} \arrayrulecolor{black}
& RLPeri  & \textbf{240.94} (2.48) & \textbf{1.448} (0.027) \\  \arrayrulecolor{black}\hline
\multirow{3}{*}{3} & ZEST  & 258.41 (0.53) & 1.897 (0.017) \\\arrayrulecolor{lightgray} \cline{2-4} \arrayrulecolor{black}
& SORS  & 279.29 (0.32) & \textbf{1.764} (0.016) \\ \arrayrulecolor{lightgray} \cline{2-4} \arrayrulecolor{black}
& RLPeri & \textbf{221.19} (3.10) & 1.906 (0.083) \\ \arrayrulecolor{black}\hline
- & PASS & 108.07 (14.76) & 15.03 (14.87) \\ \hline
- & TOP &  54.00 (0.00) & 32.35 (12.99) \\ \hline
- & DS & 156.31 (12.81) & 15.17 (11.04) \\ \hline
\end{tabular}
\caption{Performance of different perimetry strategies. Mean and standard deviation is provided for number of stimuli presented and MSE. Results are aggregated over five different runs.}
\label{tab:results}
\end{table}

We use ZEST method with random sequence of locations and random initial stimulus values as our baseline method. We also compare with SORS \citep{kucur2017sequentially}, PASS \citep{kucur2019patient}, DS \citep{weber1990topographical} and TOP strategy \citep{morales2000comparison}. SORS determines sequence of locations to be tested for reconstructing VFs more accurately and uses these reconstructed values to determine the threshold. Similar to RLPeri, it uses ZEST to estimate sensitivity threshold at a location. Unlike the SORS paper, we chose not to enforce the maximum limit of 4 stimuli per location in our implementation
to ensure fairness in comparison. PASS is an RL based method that uses a predetermined number of test locations to estimate sensitivity thresholds at different locations in the visual field. On the other hand, both DS and TOP use sensitivity threshold of tested neighbour locations as the starting stimuli value. TOP tests each location only once, yielding very fast VFs with low-accuracy. For DS and TOP, we use results presented in paper \citep{kucur2019patient}. 

Table \ref{tab:results} provides experimental results for RLPeri and other methods. $\sigma$, which is the stopping criteria for ZEST, is the predefined value of the standard deviation of the pdf. As can be seen, with increasing value of $\sigma$, MSE error increases and number of stimuli needed to complete the testing decreases. This is expected as for more accurate estimate of sensitivity threshold, a location needs to be tested with high number of stimuli. For all the values of $\sigma$, RLPeri is the fastest with comparable accuracy as compared to ZEST and SORS. Though MSE values are very close (which is expected as it is dependent on ZEST's stopping criteria), paired t-test confirmed that they are significantly different at a p-value of 1\%. This indicates that starting stimulus value and the sequence of testing locations do impact the accuracy. Overall, depending on the stopping criteria, RLPeri is faster than ZEST by around 10-15\% and faster than SORS by around 15-20\%.

TOP is a fast method as it tests each location with only one stimuli, but its accuracy is low with a high MSE of 32.35. Likewise DS is fast with 156.31 stimuli presented, it has a high MSE of 15.17. As RLPeri tests every location, for fair comparison we include results for PASS with every location tested. Although PASS is faster (108.07 number of stimuli), its MSE is high at 15.03. RLPeri's effectiveness is demonstrated by the low number of stimuli presented while maintaining the accuracy. 

\subsection{Ablation Studies}
To demonstrate the efficacy of reward shaping technique, we conduct ablation studies employing different reward functions. The outcomes of these studies are summarized in Table \ref{tab:rshaping}. The \textit{num stimuli} reward type employs the number of stimuli required to conclude the test as the reward function, the \textit{reconstruction} reward type employs the MSE as the reward function, and the \textit{shaping} reward type corresponds to the reward shaping approach detailed in previous section. Ideally, the \textit{num stimuli} reward type would prioritize speed over accuracy. Surprisingly, our observations revealed that employing the number of stimuli as a reward did not consistently result in policies that generated faster tests. This might be attributed to the fact that it overlooked the importance of recommending optimal initial stimulus values. Conversely, the \textit{reconstruction} reward type, as anticipated, led to relatively slower tests while achieving marginally enhanced accuracy. Reward shaping demonstrates consistent performance, contrasting with the other two methods, as evidenced by its reduced variance (as indicated by the standard deviation within parentheses). These results suggest that reward shaping can be leveraged to strike a balance between speed and accuracy. In the Appendix, we offer a comparative analysis by employing a 2D matrix to represent the state, in contrast to the proposed utilization of a pair of 3D matrices.

\begin{table}
\centering
\begin{tabular}{ | c | c | c | c | }
\hline 
\multirow{2}{*}{$\sigma$} & Reward & Number of & \multirow{2}{*}{MSE} \\ 
& type & stimuli presented &\\ \hline
\multirow{3}{*}{1} & num stimuli & 329.42 (33.72) & 0.952 (0.013)  \\ \arrayrulecolor{lightgray}\cline{2-4} \arrayrulecolor{black}
& reconstruction  & 347.09 (30.43) & \textbf{0.932} (0.004) \\ \arrayrulecolor{lightgray} \cline{2-4} \arrayrulecolor{black}
& shaping & \textbf{310.56} (3.2) & 0.938 (0.013)  \\  \arrayrulecolor{black}\hline
\multirow{3}{*}{2} & num stimuli  & 254.32 (26.43) & 1.482 (0.033) \\ \arrayrulecolor{lightgray} \cline{2-4} \arrayrulecolor{black}
& reconstruction  & 276.25 (35.50) & \textbf{1.442} (0.009) \\ \arrayrulecolor{lightgray} \cline{2-4} \arrayrulecolor{black}
& shaping  & \textbf{240.94} (2.48) & 1.448 (0.027) \\  \arrayrulecolor{black}\hline
\multirow{3}{*}{3} & num stimuli  & \textbf{216.82} (4.92) & 1.978 (0.166) \\\arrayrulecolor{lightgray} \cline{2-4} \arrayrulecolor{black}
& reconstruction  & 251.19 (31.77) & \textbf{1.82} (0.095) \\ \arrayrulecolor{lightgray} \cline{2-4} \arrayrulecolor{black}
& shaping & 221.19 (3.10) & 1.906 (0.083) \\ \arrayrulecolor{black}\hline
\end{tabular}
\caption{Ablation studies. Results are aggregated over five different runs.}
\label{tab:rshaping}
\end{table}

\section{Related Work}
Researchers aim to accelerate visual field testing by developing methods that efficiently manipulate stimuli values at a location. SITA (Swedish Interactive Thresholding Algorithm) is a family of visual field testing methods that uses a Bayesian adaptive approach to estimate the patient's threshold sensitivity at each location in the visual field. These algorithms also estimate the level of certainty with which the threshold is known at each point, and testing is stopped at each test location based on predetermined levels of threshold certainty. SITA standard \citep{bengtsson1997new} is the original algorithm whereas SITA Fast \citep{bengtsson1998sita} and SITA Faster \citep{heijl2019new} were developed as faster alternatives to SITA standard, with SITA Faster being the fastest but also the least accurate. To estimate threshold value at a location, ZEST \citep{king1994efficient} assigns a prior probability density function (pdf) to the visual field location, which represents the likelihood of a particular stimuli value being the sensitivity threshold of that location. In our work we use ZEST to determine the sensitivity threshold at a specific location.

Some methods find initial stimulus values by considering the interdependence of thresholds at neighboring locations, aiming to choose stimuli values that maximize information gained per stimulus. For example, DS (Dynamic Strategy) \citep{weber1990topographical} method uses the sensitivity threshold of the neighboring locations as the starting stimuli value, and adjusts the stimuli values at each location based on the response obtained at the previous location. Likewise, GOANNA (Gradient-Oriented Automated Natural Neighbor Approach) \citep{chong2014customized} uses a technique called natural neighbor interpolation to estimate the visual field sensitivity at untested locations based on the sensitivity values at tested locations. SWeLZ (Spatially Weighted Likelihoods in Zest) \citep{rubinstein2016incorporating} is a modification of the ZEST algorithm that takes into account the spatial relationships between adjacent locations in the visual field.

Efforts have been made to decrease the number of locations required to complete the visual field test in order to further expedite the process. In TOP (Tendency Oriented Perimetry) \citep{morales2000comparison}, the stimuli are presented only once at each location, and the test sequence is predetermined based on the location's tendency to have a visual field defect. SEP (Spatial entropy pursuit) \citep{wild2017spatial} utilizes entropy, which quantifies the level of uncertainty or randomness in a system, to identify the most informative locations to test and thereby minimize the number of required test locations. 

Another research direction is to find optimal sequence of these locations. SORS (Sequentially Optimized Reconstruction Strategy) \citep{kucur2017sequentially} uses an adaptive algorithm to determine the order in which visual field locations are tested to reconstruct the visual field with the greatest accuracy. Conversely, PASS (Patient-Adaptive Scheduling System) \citep{kucur2019patient} employs reinforcement learning to determine the optimal sequence of a predetermined number of test locations to estimate sensitivity thresholds. However, these methods often prioritize either speed (PASS) or accuracy (SORS). Our approach aligns with this category, utilizing reinforcement learning to determine the optimal sequence for locations during visual field testing and implementing reward shaping to strike a balance between speed and accuracy. Our experimental results demonstrate that our proposed method is faster than SORS and more accurate than PASS.

\section{Conclusion}
In this study, we introduce RLPeri, a reinforcement learning-based strategy to optimize visual perimetry testing. Our approach focuses on determining an optimal sequence of locations and initial stimulus values, aiming to reduce examination time while maintaining accuracy. Incorporating reward shaping techniques further enhances testing performance. By representing the test's state with 3D matrices and employing specialized convolutional kernels, we extract spatial and stimulus-specific features. Our experiments reveal a noteworthy 10-20\% decrease in examination time while preserving accuracy. This finding points towards a promising avenue for improving the efficiency and user-friendliness of visual perimetry testing, a critical procedure in diagnosing and monitoring conditions like glaucoma and other ocular and neurological disorders. 

\section{Acknowledgements}
This work was supported by the Agency for Science, Technology and Research (A*STAR) through its AME Programmatic Funding Scheme Under Project A20H4b0141.
\bibliography{vf}

\appendix

\onecolumn
\setcounter{secnumdepth}{1}
\graphicspath{ {./appendix-images/}}

\section*{Appendix: RLPeri: Accelerating Visual Perimetry Test with Reinforcement Learning and Convolutional Feature Extraction}

\vspace{3cm}

\section{RLPeri}
 In this section, we present a comprehensive overview of the RLPeri approach and the underlying ZEST algorithm. Algorithm \ref{alg:rlperi} outlines the detailed steps involved in learning action-values using RLPeri, which employs $\epsilon$-greedy exploration to generate learning episodes. Testing all $L$ locations based on a ground truth visual field is considered as an episode where ground truth values are used to generate \textit{seen} and \textit{not seen} responses and computing reconstruction errors. Upon selecting the next location $a_l$ and the corresponding initial stimulus value $a_v$, ZEST is employed to determine the reward $r$, representing the negative of the number of steps taken to complete the test at location $a_l$, and the estimated sensitivity threshold $v$ at that location. Reward shaping is applied based on potential values of current state and next state. Subsequently, after collecting experiences $<s, a_l, a_v, r, s'>$, the algorithm iteratively updates network parameters to minimize loss, facilitating convergence. 
\begin{algorithm} [tbhp]
\caption{RLPeri}
\label{alg:rlperi}
\textbf{Input}: Ground truth visual fields $G$, stopping standard deviation value $\sigma$.\\
\textbf{Output}: Trained model.
\begin{algorithmic}[1]
\STATE {Initialize replay buffer $\mc{J}$, action-value functions $Q_l(s,a;\theta)$, $Q_v(s,a;\theta)$ and target functions $Q_l(s,a;\theta^-)$, $Q_v(s,a;\theta^-)$}
\STATE{Initialize initial probability density function, $init\_pdf$ for every location}
\WHILE{not converged}
\FOR[Loop over ground truth VFs]{$g \in G$}
\STATE{Initialize \textit{seen} and \textit{not seen} matrices to get current state $s$}
\STATE{Initialize threshold values $pred[i] = -1, i\in L$}
\STATE{Initialize frequency of seeing $fos$ for every location based on $g$}
    \FOR[Loop over number of testing locations]{$i \in |L|$}
        \STATE{Compute potential $\varphi(s)$ based on $g$ and $pred$}
	\STATE{With probability $\epsilon$, select random action\\ 
    	$\quad a_l \leftarrow \text{random location from untested locations}$ \\
            $\quad a_v \leftarrow \text{random initial stimulus value}$ \\
    	With remaining probability $1-\epsilon$ \\
    	$\quad a_l \leftarrow  \underset{a'_l}{\mathrm{argmax}} Q_l(s, a'_l; \theta)$ \\
            $\quad a_v \leftarrow  \underset{a'_v}{\mathrm{argmax}} Q_v(s, a'_v; \theta)$ }
        \STATE{Estimate threshold using ZEST, receive corresponding reward and move to next state\\ $\quad -r, v, s' \leftarrow ZEST(a_v, init\_pdf[a_l], fos[a_l], s, \sigma)$}
        \STATE{$pred[a_l]=v$}
        \STATE{Compute potential $\varphi(s')$ based on $g$ and $pred$}
        \STATE{$r \leftarrow r + \gamma \varphi(s') - \varphi(s)$}
        \STATE{$s \leftarrow s'$}
        \STATE{Store experience $<s, a_l, a_v, r, s'>$ in the replay memory $\mc{J}$}
    \ENDFOR
\STATE{Periodically update the network parameters $\theta$ by sampling experiences from $\mc{J}$ and minimizing the loss function  $ \mc{L} = \bb{E}_{(s,a_l, a_v,r,s') \sim \mc{J}} \Big[\dfrac{1}{2}\big[ \big(y - Q_l(s,a_l; \theta)\big) + \big(y - Q_v(s,a_v; \theta)\big) \big] \Big]^2 $}

\STATE{Periodically update the target network parameters}
\ENDFOR
\ENDWHILE
\STATE {\textbf{return} $Q_l(s,a;\theta)$, $Q_v(s,a;\theta)$}
\end{algorithmic}
\end{algorithm} 

The details of ZEST is provided in Algorith \ref{alg:zest}. It estimates threshold values during perimetry testing. It begins with an initial stimulus value and iteratively adjusts the probability density function based on patient responses and likelihood. The algorithm stops when stopping criterion is met. It returns the number of stimuli presented, estimated threshold value, and the next state.

\begin{algorithm}[tbhp]
\caption{ZEST}
\label{alg:zest}
\textbf{Input}: Initial stimulus value $a_v$, initial pdf $init\_pdf$, probability of seeing $p$, current state $s$, stopping criteria $\sigma$.\\
\textbf{Output}: Number of stimuli presented $n$, estimated threshold value $v$, next state $s'$.
\begin{algorithmic}[1] 
\STATE $n=0, v=a_v, min=0, max=40$
\STATE $vals=[min,...,max]$
\STATE $terminate=False$
\WHILE{$!terminate$ and $v\geq min$ and $v \leq max$}
    \STATE $rand=U(0,1)$ \COMMENT{sample from uniform distribution}
    \STATE Determine patient's response, $resp=(p > rand)$
    \STATE Update state to $s'$ based on response 
    \STATE Get likelihood function, $l=likelihood(resp,v, vals)$
    \STATE Update the pdf estimate, $pdf=pdf*l$
    \STATE $pdf=norm(pdf)$
    \STATE $terminate=std(pdf)<\sigma$
    \STATE $v=argmax(pdf)$
\ENDWHILE
\STATE \textbf{return} $n,v,s'$
\end{algorithmic}
\begin{algorithmic}[1]
\STATE \textbf{Function} $likelihood$ ($resp,v, vals$):
    \STATE $l=Gaussian\_cdf(vals,mean=v,std=1)$
    \STATE $l=resp*(resp-l)+(1-resp)*l$
\STATE \textbf{return $l$}
\end{algorithmic}
\end{algorithm}

\vspace{2cm}

\section{Ablation Studies}
In this section, we present a comparison of results obtained when utilizing a 2D matrix to represent the state, as opposed to the proposed approach using a pair of \textit{seen} and \textit{not seen} matrices. Additionally, we offer a qualitative analysis of the generated sequence of locations and initial stimulus values across multiple visual fields.
\vspace{1cm}
\subsection{State Representation}
To assess the effectiveness of the proposed state representation using a pair of 3D matrices, we conduct a comparison with an alternative state representation. The alternative approach involves representing the state as a 2D matrix, where visual locations are mapped onto the matrix and estimated threshold values are assigned as the entries. We used a $8 \times 9$ matrix to represent the 54 locations, masking the corner locations that do not belong to the visual field. Specifically, the non-visual field locations were assigned a constant value of -2. For the remaining locations, we initialized the threshold sensitivity values to values outside the valid range (-1), which were subsequently updated withe estimated threshold values during testing.

Table \ref{tab:state} shows the results for different $\sigma$ values. During training, reward shaping was employed for both type of state representation. The number of stimuli presented (indicative of speed) and MSE (accuracy) values both exhibit inferior performance with the 2D state representation. Moreover, the high variance indicates inconsistent performance when using the 2D matrix state representation. This observation supports our assertion that representing the state as a pair of \textit{seen} and \textit{not seen} matrices leads to more efficient extraction of state features, resulting in the learning of improved policies.
\begin{table} [h]
\centering
\caption{Ablation studies. Mean and standard deviation is provided for number of stimuli presented and MSE. Results are aggregated over five different runs.}
\begin{tabular}{ | c | c | c | c | }
\hline 
\multirow{2}{*}{$\sigma$} & State & Number of & \multirow{2}{*}{MSE} \\ 
& type & stimuli presented &\\ \hline
\multirow{3}{*}{1} & 2D state & 328.99 (14.26) & 0.952 (0.008)  \\ \arrayrulecolor{lightgray}\cline{2-4} \arrayrulecolor{black}
& 3D state & 310.56 (3.2) & 0.938 (0.013)  \\  \arrayrulecolor{black}\hline
\multirow{3}{*}{2} & 2D state  & 257.41 (10.9) & 1.486 (0.009) \\ \arrayrulecolor{lightgray} \cline{2-4} \arrayrulecolor{black}
& 3D state  & 240.94 (2.48) & 1.448 (0.027) \\  \arrayrulecolor{black}\hline
\multirow{3}{*}{3} & 2D state  & 235.67 (21.42) & 1.965 (0.091) \\\arrayrulecolor{lightgray} \cline{2-4} \arrayrulecolor{black}
& 3D state & 221.19 (3.10) & 1.906 (0.083) \\ \arrayrulecolor{black}\hline
\end{tabular}
\label{tab:state}
\end{table}
\vspace{1cm}
\subsection{Qualitative Analysis}
\begin{figure*}[h!tb]
  \centering
    \subfloat[Ground truth A \label{fig:gta}]{\includegraphics[scale=0.4]{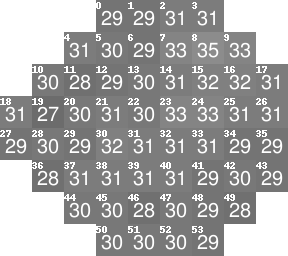}} \qquad
     \subfloat[Ground truth B \label{fig:gtb}]{\includegraphics[scale=0.4]{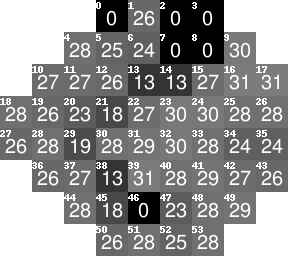}} \qquad
     \subfloat[Ground truth C \label{fig:gtc}]{\includegraphics[scale=0.4]{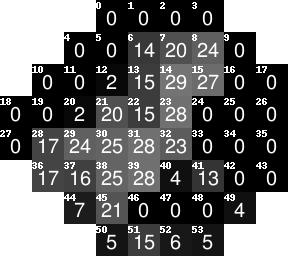}} \\
     \subfloat[Reconstructed VF A \label{fig:finala}]{\includegraphics[scale=0.4]{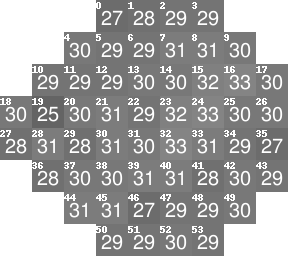}} \qquad
     \subfloat[Reconstructed VF B \label{fig:finalb}]{\includegraphics[scale=0.4]{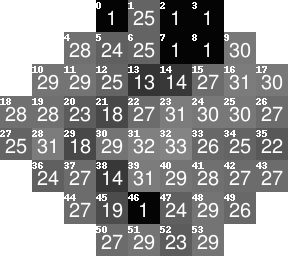}} \qquad
     \subfloat[Reconstructed VF C \label{fig:finalc}]{\includegraphics[scale=0.4]{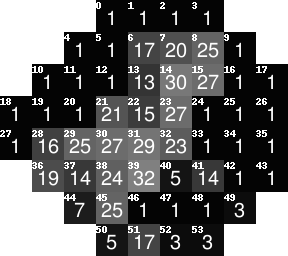}} \\
     \subfloat[Initial stimulus value A \label{fig:inita}]{\includegraphics[scale=0.4]{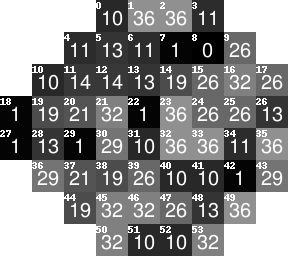}} \qquad
     \subfloat[Initial stimulus value B \label{fig:initb}]{\includegraphics[scale=0.4]{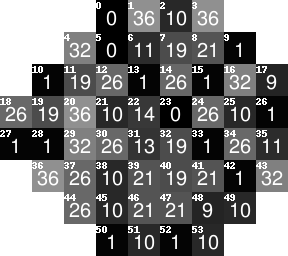}} \qquad
     \subfloat[Initial stimulus value C \label{fig:initc}]{\includegraphics[scale=0.4]{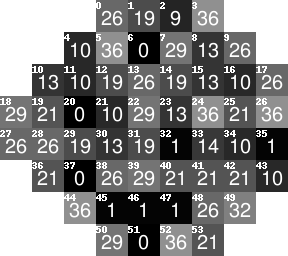}} \\
     \subfloat[Sequence of locations A \label{fig:indexa}]{\includegraphics[scale=0.4]{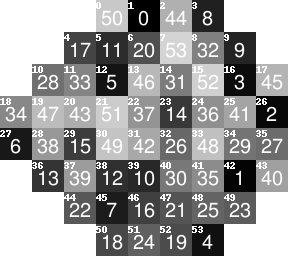}} \qquad
     \subfloat[Sequence of locations B \label{fig:indexb}]{\includegraphics[scale=0.4]{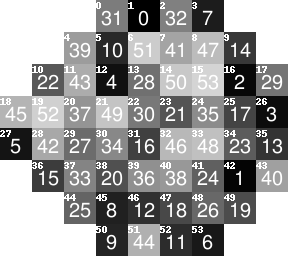}} \qquad
     \subfloat[Sequence of locations C\label{fig:indexc}]{\includegraphics[scale=0.4]{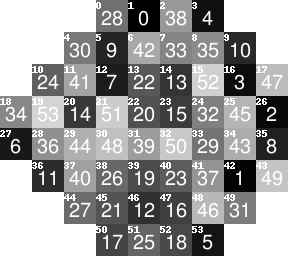}} \\
     \subfloat[Number of stimuli A \label{fig:rwa}]{\includegraphics[scale=0.4]{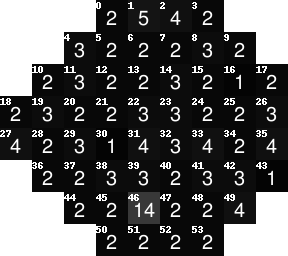}} \qquad
     \subfloat[Number of stimuli B \label{fig:rwb}]{\includegraphics[scale=0.4]{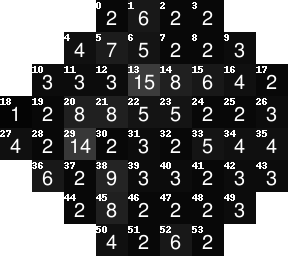}} \qquad
     \subfloat[Number of stimuli C \label{fig:rwc}]{\includegraphics[scale=0.4]{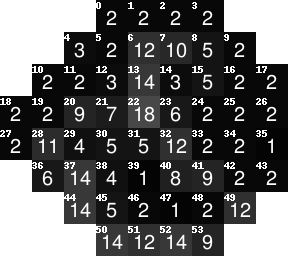}} 
  \caption{Ground truth values, reconstructed visual fields, initial stimulus presented, sequence of locations and number of stimuli presented at each location for testing of three different visual fields A, B and C.}
  \label{fig:steps}
\end{figure*}

Figure \ref{fig:steps} provides a qualitative understanding of RLPeri by displaying ground truth values, reconstructed VFs, initial stimulus values, the sequence of testing locations, and the number of stimuli presented at each location for different visual fields. We selected three visual fields with varying degree of sensitivities and we used stopping criterion $\sigma=3$. The ground truth VFs are represented in Figures \ref{fig:gta}-\ref{fig:gtc} with darker shades indicating lower sensitivity. The corresponding reconstructed VFs are displayed in Figures \ref{fig:finala}-\ref{fig:finalc}. The sequence of testing locations used is depicted in Figures \ref{fig:indexa}-\ref{fig:indexc}, with darker shades indicating early tested locations. Figures \ref{fig:rwa}-\ref{fig:rwc} illustrate the number of stimuli presented (following the ZEST algorithm) at each location. The total number of stimuli presented were 146 for VF A, 219 for VF B, and 300 for VF C. The reconstruction MSE errors were 1.56, 1.80 and 2.07 respectively. It is interesting  to note that due to probabilistic responses, presenting initial stimulus values closer to the ground truth value does not necessarily result in fewer stimuli presented to estimate the threshold. For instance, for VF A, even if the initial stimulus value of 32 (Figure \ref{fig:inita}) for location 46 is close to the ground truth value of 28 (Figure \ref{fig:gta}), it still required 14 stimuli (Figure \ref{fig:rwa}) to estimate the threshold. On the contrary, even if the initial stimulus value of 26 (Figure \ref{fig:initc}) at location 0 (Figure \ref{fig:indexc}) is far from the ground truth value of 0 (Figure \ref{fig:gtc}), it only took 2 stimuli (Figure \ref{fig:rwc}) to estimate the threshold as 1 (Figure \ref{fig:finalc}). Figures \ref{fig:indexa}-\ref{fig:indexc} demonstrate that the sequence of initial tested locations (darker shade locations) are similar when a test is started, however the sequence changes as the test proceeds and threshold values are estimated. This observation indicates that the algorithm has learned to suggest the next location based on the current state. Similarly, the recommendation for initial stimulus value also improves as testing progresses. 

\end{document}